\documentclass[11pt]{article}

\usepackage[utf8]{inputenc}
\usepackage[T1]{fontenc}
\usepackage{newtxtext,newtxmath}
\usepackage{microtype}
\usepackage[margin=1in]{geometry}

\usepackage{amsmath}

\usepackage{graphicx}
\usepackage{booktabs}
\usepackage{array}
\usepackage{multirow}
\usepackage[table]{xcolor}
\usepackage{subcaption}
\usepackage{float}
\usepackage{wrapfig}
\usepackage{placeins}

\usepackage{algorithm}
\usepackage{algorithmic}

\usepackage[numbers,sort&compress]{natbib}

\usepackage[font=small,labelfont=bf]{caption}
\usepackage{enumitem}

\usepackage{hyperref}
\hypersetup{colorlinks=true,linkcolor=blue,citecolor=blue,urlcolor=blue}
\usepackage{url}

\title{Complementarity-Supervised Spectral-Band Routing for Multimodal Emotion Recognition}

\author{%
  Zhexian Huang\textsuperscript{1,2,*}\quad
  Bo Zhao\textsuperscript{1,2,*}\quad
  Hui Ma\textsuperscript{1,\dag}\quad
  Zhishu Liu\textsuperscript{1,3}\quad
  Jie Zhang\textsuperscript{1}\\[0.3em]
  Ruixin Zhang\textsuperscript{4}\quad
  Shouhong Ding\textsuperscript{4}\quad
  Zitong Yu\textsuperscript{1,\dag}\\[0.8em]
  \textsuperscript{1}Great Bay University\quad
  \textsuperscript{2}The Chinese University of Hong Kong, Shenzhen\\
  \textsuperscript{3}City University of Hong Kong (Dongguan)\quad
  \textsuperscript{4}Tencent Youtu Lab\\
  \texttt{225040199@link.cuhk.edu.cn}\quad
  \texttt{yuzitong@gbu.edu.cn}\\[0.3em]
  {\small \textsuperscript{*}Equal contribution\quad \textsuperscript{\dag}Corresponding authors}
}

\date{}
\begin{document}
\maketitle

\begin{abstract}
Multimodal emotion recognition fuses cues such as text, video, and audio to understand individual emotional states. Prior methods face two main limitations: mechanically relying on independent unimodal performance, thereby missing genuine complementary contributions, and coarse-grained fusion conflicting with the fine-grained representations required by emotion tasks. As inconsistent information density across heterogeneous modalities hinders inter-modal feature mining, we propose the Complementarity-Supervised Multi-Band Expert Network, named Atsuko, to model fine-grained complementary features via multi-scale band decomposition and expert collaboration. Specifically, we orthogonally decompose each modality's features into high, mid, and low-frequency components. Building upon this band-level routing, we design a modality-level router with a dual-path mechanism for fine-grained cross-band selection and cross-modal fusion. To mitigate shortcut learning from dominant modalities, we propose the Marginal Complementarity Module (MCM) to quantify performance loss when removing each modality via bi-modal comparison. The resulting complementarity distribution provides soft supervision, guiding the router to focus on modalities contributing unique information gains. Extensive experiments show our method achieves superior performance on the CMU-MOSI, CMU-MOSEI, CH-SIMS, CH-SIMSv2, and MIntRec benchmarks. Our code is publicly available at: https://github.com/EddyCodex/Atsuko
\end{abstract}

\section{Introduction}
\label{sec:intro}
Emotion is closely related to human cognition, decision-making, and behavior, playing a crucial role in daily life~\cite{Picard2000Affective}. Since human behavior conveys emotions in various forms, multimodal emotion recognition aims to predict human emotions by leveraging multiple modalities, including visual, acoustic, and linguistic information. Compared to unimodal methods, distinct modalities typically offer complementary, non-redundant information~\cite{Wei2024DMRNet,Wei2024DiagRelearn,Ye2025CATPlus}, enhancing affective behavior prediction and aligning more closely with natural human emotional expression.

In principle, linguistic content, facial expressions, and acoustic prosody provide complementary cues to jointly distinguish subtle emotional states. However, in real-world segments, modality informativeness and reliability vary across samples. Fixed fusion paradigms fail to capture sample-dependent modality importance, potentially biasing learning toward dominant modalities~\cite{Wei2024DiagRelearn,Wei2024DMRNet}. Furthermore, many fusion frameworks overemphasize cross-modal invariants, inadvertently suppressing modality-specific cues and losing specificity~\cite{Fang2025EMOE}. These issues are especially pronounced in MER, where individuals display inconsistent cross-modal expressions and cue availability varies drastically across segments.

\begin{figure}[t]
    \centering
    \includegraphics[width=0.9\linewidth]{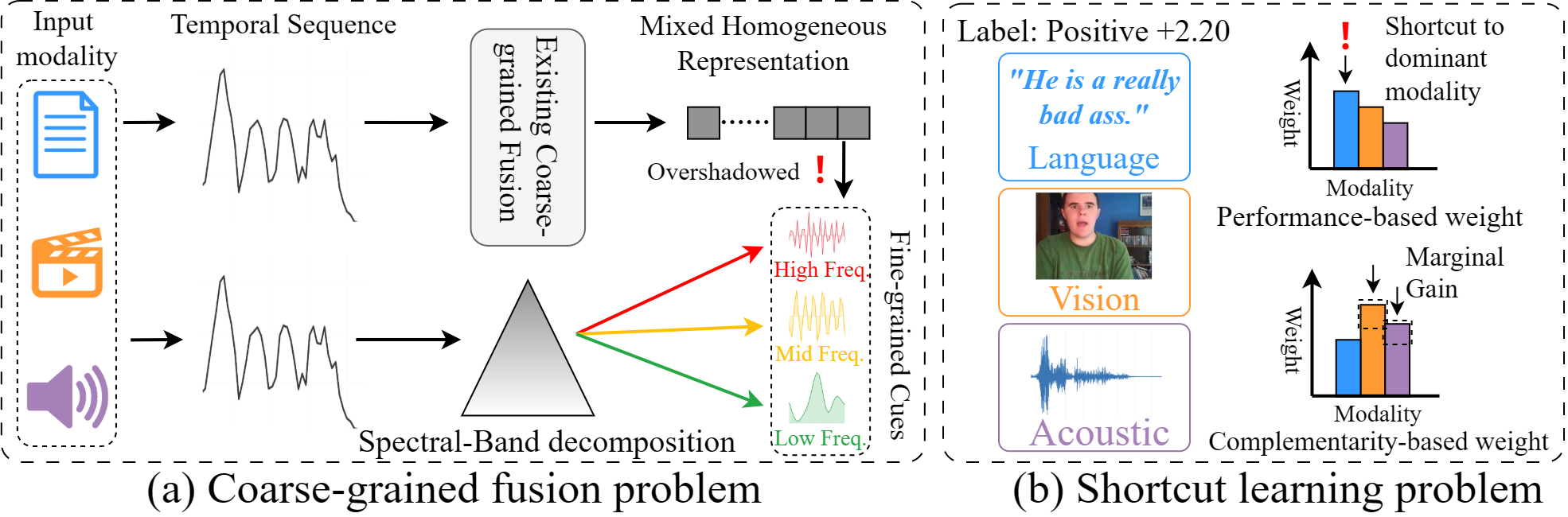}
    \caption{\textbf{Two fundamental limitations of existing MER methods.}
    \textbf{(a)} Coarse-grained fusion: treating each modality's temporal features as 
    a homogeneous signal discards fine-grained emotional cues in specific frequency bands.
    \textbf{(b)} Shortcut learning: performance-based weighting over-relies on the 
    dominant (text) modality, whereas visual and acoustic modalities carry the genuinely complementary cues. Complementarity-based weighting correctly amplifies these contributions.}
    \label{fig:motivation}
\end{figure}

Substantial research has advanced across cross-modal fusion, modality disentanglement, and dynamic modal balancing. Traditional paradigms evolved from early and late fusion to modeling high-order interactions~\citep{Zadeh2017TFN,Liu2018LMF} and unaligned cross-modal attention architectures~\citep{Tsai2019MulT,Zadeh2018MFN,Yuan2024AUFormer}. Despite enhancing interactions, these methods impose uniform fusion rules across samples, ignoring the strong sample-level heterogeneity of emotional expressions. Parallel disentanglement approaches separate shared invariant from private specific subspaces~\citep{Hazarika2020MISA} or perform heterogeneous cross-modal distillation~\citep{Wei2024DMRNet} to preserve modality-specific cues, generally lacking direct constraints on individual contributions at the final decision stage. Recent balancing strategies~\citep{Peng2022OGM,Wei2024DiagRelearn,Yuan2024AUFormer,Lin2025MMDGPlus} mitigate excessive reliance on dominant modalities. However, deriving learning signals from unimodal separability, training dynamics, or heuristic proxy metrics inaccurately reflects the genuine marginal, non-redundant information gain each modality contributes to the multimodal system. As illustrated in Figure~\ref{fig:motivation}(b), a modality with strong individual performance may contribute limited incremental information when integrated, whereas a seemingly weaker modality can provide critical complementary cues that significantly enhance multimodal prediction~\citep{Peng2023CARAT,Zhang2024MLA}.

Beyond modality-level heterogeneity, MER signals exhibit structured intra-modal variability~\citep{Wei2024DMRNet,Wei2024DiagRelearn,Chen2023M3Net}. Emotional cues concentrate at distinct temporal scales, including subtle high-frequency facial micro-movements and prosodic fluctuations~\citep{Fan2023SelfME,Satt2017InterspeechSpec,Zou2022ICASSP,Liu2025AULLM}, and are susceptible to selective corruption by noise such as motion blur, occlusions, and background audio~\citep{Liu2024Norface,Liu2023MLKD,Yu2023ViTMAE_FAS}. Consequently, spectral analysis naturally characterizes these multiscale cues. While widely utilized in speech emotion recognition to isolate emotion-relevant frequencies~\citep{Satt2017InterspeechSpec,Luengo2009Interspeech,Zou2022ICASSP} and recently highlighted for preserving valuable high-frequency components~\citep{Chen2023M3Net}, its application to multimodal sequences remains largely unexplored. As Figure~\ref{fig:motivation}(a) illustrates, existing fusion frameworks persist in coarse-grained modality-level designs, typically treating temporal features as homogeneous signals and neglecting that distinct frequency components carry varying emotional information.

To address these challenges, we propose the Complementarity-Supervised Multi-Band Expert Network, named Atsuko, to extract fine-grained complementary features via spectral decomposition and expert collaboration. First, to overcome inconsistent information density and selectively attend to emotionally relevant frequencies, we introduce a spectral-band hierarchical routing mechanism. It orthogonally decomposes temporal signals into distinct frequency bands via graph Laplacian eigenvectors and applies a data-driven equal-energy criterion. A lightweight band router then dynamically weights these components based on sample characteristics. Subsequently, a modality-level router with a dual-path decision mechanism recombines these band-filtered features for cross-modal fusion, achieving granular control from spectral details to global representations. Second, to mitigate shortcut learning from dominant modalities, we propose the Marginal Complementarity Module (MCM) inspired by the Shapley value. By evaluating bimodal scenarios where individual modalities are absent, MCM quantifies each modality's marginal performance contribution relative to the full trimodal prediction. The resulting complementarity distribution provides soft supervision via Kullback-Leibler divergence minimization to guide the modality router, ensuring the fusion prioritizes modalities supplying unique information gains over redundant independent performance.

The main contributions of this paper are summarized as follows:
\begin{itemize}
    \item We propose the Spectral-Band Hierarchical Routing Network, which uses a lightweight router to dynamically weight frequency bands based on the signal-to-noise ratio and incorporates a random masking mechanism, effectively enhancing the capability of the model to capture fine-grained high-frequency cues and its robustness against noise interference.
    \item We design a modality-level router equipped with a dual-path decision mechanism that supports flexible switching between shallow feature spaces and deep semantic decision spaces, thereby achieving granular and interpretable control over the multimodal signal selection process.
    \item We introduce the Marginal Complementarity Module (MCM), employing complementarity distribution as a soft supervision signal to guide routing learning. This ensures the model focuses on modalities yielding unique information gains, thereby avoiding shortcut learning.
\end{itemize}

\section{Related Works}
\label{sec:related}
\subsection{Multimodal Emotion Recognition}
Multimodal Emotion Recognition (MER) integrates linguistic, visual, and acoustic cues to comprehend human emotions. As a pivotal artificial intelligence domain, MER underpins diverse applications, including driver emotion monitoring~\citep{Xiang2025DriverMER}, mental health screening~\citep{Ringeval2019AVEC}, and online education~\citep{ElMaazouzi2025Elearning}. Early methods focus on feature fusion via tensors~\citep{Zadeh2017TFN, Liu2018LMF} or crossmodal attention~\citep{Tsai2019MulT, Zadeh2018MFN}. However, these methods typically assume consistent modal contributions across samples, neglecting real-world modal heterogeneity and imbalance. To address this, optimization methods modulate gradient updates~\citep{Peng2022OGM} or penalize reliance on dominant modalities~\citep{Yerramilli2024AttrReg}. Yet, such global adjustments struggle to capture sample level dynamics. Recently, Shapley value approaches~\citep{Parcalabescu2022MMSHAP, SHAPE2022} have quantified marginal modal contributions for post hoc interpretability. Unlike these primarily post-hoc explanation approaches, our proposed MCM embeds the Shapley value principle directly into training to guide routing, prioritizing modalities that offer genuine complementary information gains.
Mixture of Experts (MoE) architectures have extended from unimodal tasks~\citep{Fedus2022Switch, Lepikhin2020GShard, Riquelme2021VMoE} to Large Multimodal Models~\citep{UniMoE2024, OmniSMoLA2024}. In affective computing, EMOE~\citep{Fang2025EMOE} pioneers dynamic routing via modality experts to address sample dependencies. Furthermore, hierarchical MoE frameworks~\citep{Zhu2025HiMoE} enhance robustness against modality missingness, while sparse MoE with interactive distillation~\citep{Li2025SUMMER} utilizes unimodal teachers to mitigate learning ambiguity. Although improving routing flexibility, these decisions predominantly rely on representation strength or similarity, failing to explicitly model complementary contributions. This risks excessive reliance on superficially dominant modalities. Our MCM rectifies this by quantifying the marginal contribution of each modality, explicitly discovering and exploiting multimodal complementary synergies.

\subsection{Hierarchical Temporal-Spectral Representation Learning}
Emotional cues within temporal signals inherently exhibit multiscale characteristics, where high-frequency components encode transient micro-expressions and low-frequency components convey stable emotional baselines~\citep{Chen2023M3Net, Satt2017InterspeechSpec}. Early methods leverage wavelets or Fourier spectra to extract energy from fixed bands~\citep{Luengo2009Interspeech,Zou2022ICASSP}, but static filtering fails to adapt to sample-specific signal-to-noise ratio fluctuations. Although global Fourier mixing~\citep{LeeThorp2022FNet, Rao2021GFNet} reduces attention complexity, its implicit modeling lacks interpretable band separation.
Recently, Graph Signal Processing (GSP) provides a principled framework for multimodal temporal analysis. Zhang et al.~\citep{Zhang2024GSMCC} apply graph Fourier operators to decouple low-frequency consistent semantics from high-frequency complementary ones. Meanwhile, Chen et al.~\citep{Chen2023M3Net} employ message passing on multi-frequency graphs to capture long-range dependencies and fine-grained temporal structures. In parallel, FusionMamba~\citep{Xie2024FusionMamba} integrates dynamic convolution with state-space models for efficient cross-modal feature enhancement. Additionally, AdaGNN~\citep{AdaGNN2021} and NFGNN~\citep{NFGNN2024} utilize learnable cross-layer and node-level filters to enhance spectral expressiveness.
Despite these advancements, existing frequency-domain techniques largely rely on rigid configurations such as equidistant partitioning. This inflexibility exacerbates mismatches caused by divergent cross-modal spectral energy distributions~\citep{Zhou2024FPro,Chen2024FADC,Wang2024SelectiveStereo}. Furthermore, current GSP models primarily learn static filter parameters~\citep{Defferrard2016ChebNet,He2021BernNet,Xu2024SLOG}. Applying uniform weights across frequency bands ignores sample-specific variations, often suppressing crucial high-frequency details. Consequently, the dynamic assessment of frequency band reliability tailored to local content remains unexplored in multimodal emotion recognition.
To address these gaps, we propose extending dynamic routing to the spectral domain. By utilizing equal-energy partitioning and sample-specific band weighting with random masking, our approach achieves noise-robust, fine-grained control across both frequency and modality levels.

\section{The Proposed Method}
\label{sec:method}

\begin{figure}[t]
    \centering
    \includegraphics[width=\linewidth]{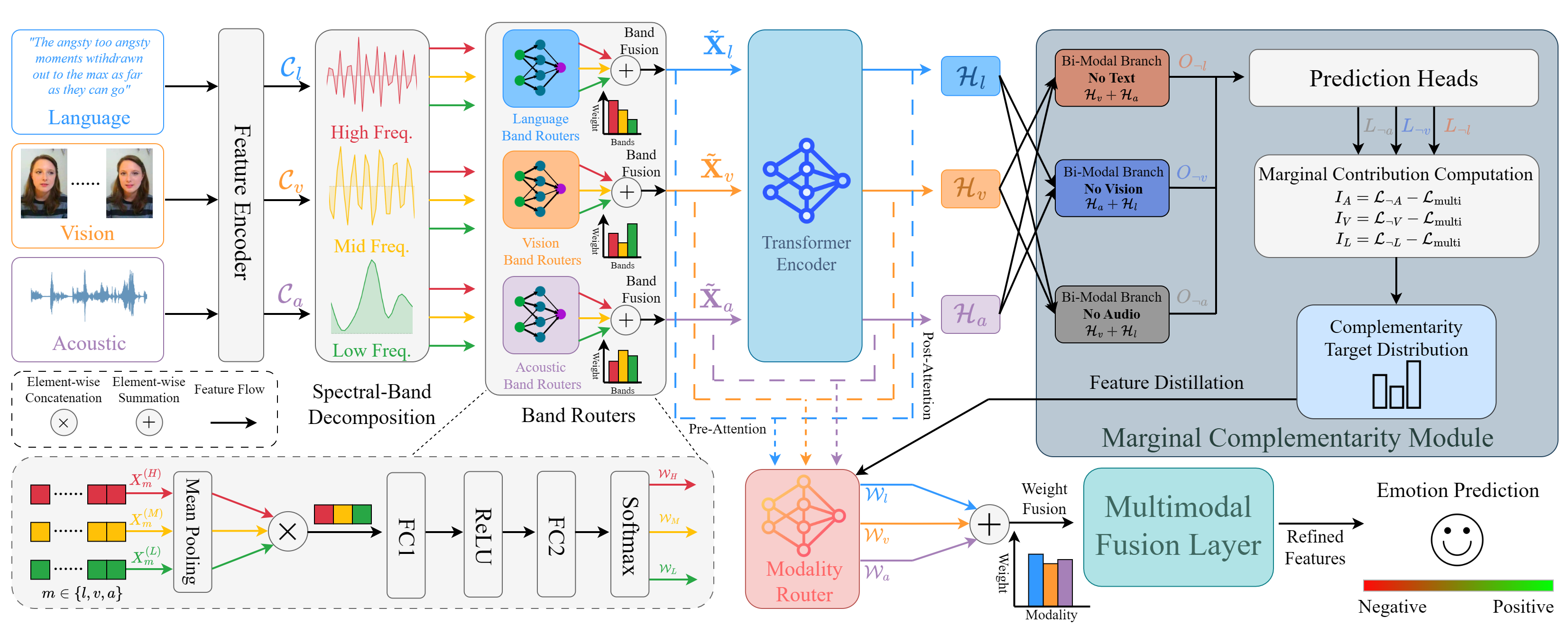}
    \caption{\textbf{The overall architecture of Atsuko.}
    \textbf{(1)} \textbf{Spectral-Band Routing:} Multi-modal temporal inputs are orthogonally decomposed into distinct frequency bands via graph Laplacian and dynamically weighted to yield band-enhanced features $\tilde{X}_m$.
    \textbf{(2)} \textbf{Modality-Level Routing:} A dual-path mechanism (pre- and post-attention) assigns dynamic fusion weights $w_m$ to the deep semantic features.
    \textbf{(3)} \textbf{Marginal Complementarity Module (MCM):} Bimodal contrastive branches quantify each modality's marginal contribution, providing soft supervision to guide the modality router toward complementary information gains, augmented by feature distillation.}
\label{fig:overview}
\end{figure}

Figure~\ref{fig:overview} illustrates the overall architecture of Atsuko. Given linguistic, visual, and acoustic inputs, the model extracts temporal representations via feature encoders and feeds them into the Spectral-Band Hierarchical Routing Network (Section~\ref{sec:sb_routing}) for band-level filtering and modality-level aggregation. The Marginal Complementarity Module (Section~\ref{sec:mcm}) then quantifies marginal contributions to guide the routing process toward complementary information gains. Finally, the fused features pass through a prediction layer to yield emotion predictions. The detailed training procedure is summarized in Alg.~1 of the supplementary material.

\subsection{Spectral-Band Hierarchical Routing Network}
\label{sec:sb_routing}
Since emotional cues are distributed across multi-scale frequency components and subject to band-specific noise, we decompose temporal signals into orthogonal spectral bands before adaptive routing.
We consider three modalities: linguistic (L), visual (V), and acoustic (A).
First, independent one-dimensional temporal convolutional layers are employed for each modality to project raw features into a unified $d$-dimensional space. Subsequently, a parameter-shared $1 \times 1$ convolutional layer is applied to perform cross-modal consistent feature transformation, yielding the temporal representations $X_m \in \mathbb{R}^{T_m \times d}$, where $m \in \{l, v, a\}$.

Unlike FFT, whose fixed sinusoidal bases assume infinite periodicity, ignoring finite-length temporal adjacency, graph Laplacian eigen-decomposition derives orthogonal bases directly from data topology~\citep{Shuman2013GSP,Defferrard2016ChebNet}. Thus, we model $X_m$ as a Temporal Path Graph $\mathcal{G}=(\mathcal{V}, \mathcal{E})$, where $\mathcal{V}$ denotes the set of $T$ time steps and $\mathcal{E}$ encompasses all adjacent connections $(t, t+1)$. Its normalized Laplacian is:
\begin{equation}
    \widetilde{L} = I - D^{-1/2} A D^{-1/2}
\end{equation}
where $A$ and $D$ are the adjacency and degree matrices, respectively. Eigen-decomposition yields $\widetilde{L} = U \Lambda U^\top$, where $U = [u_1, u_2, \dots, u_T]$ forms the orthogonal eigenvector basis. Smaller eigenvalues $\lambda_k$ correspond to smoother (low-frequency) basis vectors, while larger ones capture rapid oscillations (high frequencies). We project features into the spectral domain via $\hat{X}_m = U^\top X_m$.

To aggregate the $T$ frequency components into $K$ sub-bands, we must address the long-tail energy distribution: low-frequency components concentrate most energy, whereas high-frequency components, despite low energy, often carry transient emotional cues. Simple equidistant partitioning causes the router to neglect high-frequency bands. We therefore propose an Equal-Energy Partitioning strategy: we compute the cumulative energy distribution $\text{CDF}(f)$ by aggregating the average spectral energy density $E_f = \mathbb{E}[|\hat{X}_{m}^{(f)}|_2^2]$ over the training set and determine band boundaries $\mathcal{B} = \{b_0, b_1, \dots, b_K\}$ such that each band contains approximately equal signal energy.
Let $U^{(k)} = [u_{b_{k-1}}, \dots, u_{b_k - 1}] \in \mathbb{R}^{T \times (b_k - b_{k-1})}$ and $\hat{X}_m^{(k)} \in \mathbb{R}^{(b_k - b_{k-1}) \times d}$ denote the eigenvector sub-matrix and spectral coefficients for the $k$-th band. The band component is reconstructed via
$X_m^{(k)} = U^{(k)} \hat{X}_m^{(k)}$.
This decomposition guarantees $\sum_{k=1}^K X_m^{(k)} = X_m$, i.e., orthogonal decoupling with zero information loss. The resulting orthogonality enables the subsequent band routing to independently evaluate and weight each frequency component, preventing the systematic underestimation of high-frequency emotional cues.

To dynamically evaluate the emotional relevance of each band under sample-specific signal-to-noise variations, we design a lightweight Band Router $\mathcal{R}_{band}$. For each band component $X_m^{(k)}$, average pooling extracts the summary vector $h_m^{(k)} = \text{Pool}(X_m^{(k)}) \in \mathbb{R}^d$. All summaries are concatenated and fed into an MLP to produce dynamic weights, then the enhanced representation is refined:
\begin{equation}
    \alpha_{m} = \text{Softmax}\left(\frac{\text{MLP}([h_m^{(1)} \oplus h_m^{(2)} \oplus \dots \oplus h_m^{(K)}])}{\tau_1}\right),
    \quad \tilde{X}_m = \sum_{k=1}^K \alpha_{m, k} \cdot X_m^{(k)}
\end{equation}
where $\oplus$ denotes concatenation and $\tau_1$ is the temperature coefficient.

To further strengthen robustness, we apply Stochastic Band Masking during training: a Bernoulli mask $M \in \{0, 1\}^K$ with probability $p$ is applied to the weights as $\hat{\alpha}_{m} = (\alpha_{m} \odot M) / (\sum (\alpha_{m} \odot M) + \epsilon)$, compelling the router to seek alternative cues when specific bands are absent.

Following intra-modal band enhancement, a modality-level router
$\mathcal{R}_{modal}$ dynamically allocates fusion weights across modalities.
It is implemented as a two-layer linear network with intermediate $L_2$
normalization onto a unit hypersphere, ensuring decisions depend on feature
directionality rather than magnitude, as heterogeneous modalities inherently
produce features at disparate scales:

\begin{equation}
    w = \text{Softmax}\left(\frac{W_2 \cdot \text{ReLU}\!\big(\hat{z}\big)}{{\tau_2}}\right), \quad \hat{z} = \frac{W_1 Z_{decision}}{\|W_1 Z_{decision}\|_2}
\end{equation}
where $W_1 \in \mathbb{R}^{d_h \times d_{in}}$ and $W_2 \in \mathbb{R}^{3 \times d_h}$ are learnable parameters. Two routing pathways are supported:
\begin{enumerate}
    \item \textbf{Pre-Attention Path:} $Z_{decision} = [\text{Pool}(\tilde{X}_l) \oplus \text{Pool}(\tilde{X}_v) \oplus \text{Pool}(\tilde{X}_a)]$, computed before Transformer encoding. This decouples routing from deep feature extraction, preventing encoder output variance from interfering with weight estimation.
    \item \textbf{Post-Attention Path:} $Z_{decision} = [H_l \oplus H_v \oplus H_a]$, where $H_m = \mathcal{F}_{enc}(\tilde{X}_m)_{[-1]}$ is the final-position output of the Transformer encoder. Self-attention implicitly aggregates the full sequence context, providing a compact semantic summary suitable for subtle emotional expressions requiring contextual reasoning.
\end{enumerate}
The deep semantic representations are then weighted and summed, and then refined via a residual MLP. The prediction head then outputs the emotion prediction.
\begin{equation}
O_{multi} = \sum_{m} w_m H_m,
\quad
O'_{multi} = \mathrm{MLP}_{multi}(O_{multi}) + O_{multi}
\end{equation}

\subsection{Marginal Complementarity Module}
\label{sec:mcm}
The fundamental challenge in multimodal fusion lies in ensuring that each modality contributes unique, irreplaceable information rather than redundant common features. A strong modality may exhibit substantial redundancy with others, whereas a seemingly weaker modality can offer critical information gain. To capture this, we propose the Marginal Complementarity Module (MCM), which quantifies the marginal contribution of each modality $m$ by comparing system performance with and without it, rather than relying on standalone unimodal performance.
We construct three parallel bimodal branches that simulate scenarios where each modality is absent. For modality $m$, the bimodal prediction is computed from the remaining deep semantic features $H_{m'}$ ($m' \neq m$), reusing the shared semantic encoders with independent prediction heads:
\begin{align}
    O_{\neg m} = \mathcal{F}_{\neg m}\left(\sum_{m' \neq m} H_{m'}\right), \quad m \in \{l, v, a\}
\end{align}
where $\mathcal{F}_{\neg m}$ is a prediction network for the specific bimodal combination. Jointly trained with the full-modal branch to approximate the ground truth $y$, these branches establish reliable baselines whose performance gap with $O_{multi}$ reflects the excluded modality's marginal effect.
We define Complementarity Importance $I_m$ as the per-sample difference between bimodal and full-modal losses, applying a stop-gradient operation to prevent branch training interference. The raw scores are then normalized into the complementarity target distribution:
\begin{equation}
    I_m = \text{StopGradient}(\mathcal{L}_{\neg m} - \mathcal{L}_{multi}), \quad P_{comp} = \text{Softmax}([I_l, I_v, I_a] / \tau_{comp})
\end{equation}
where $\tau_{comp}$ controls the distribution sharpness. We inject this complementarity prior into the routing process by minimizing the KL divergence between $P_{comp}$ and the modality routing weights $w$ (Section~\ref{sec:sb_routing}):
\begin{equation}
    \mathcal{L}_{mcm} = D_{KL}(P_{comp} \parallel w) = \sum_{m \in \{l,v,a\}} P_{comp}[m] \log \left( \frac{P_{comp}[m]}{w_m} \right)
\end{equation}
This constraint guides the router to prioritize modalities yielding genuine complementary gains, circumventing shortcut learning from dominant modalities. To ensure the fused representation preserves router-identified complementary information, we introduce feature distillation. Each modality's semantic representation is refined via an independent residual MLP: $H_m' = \text{MLP}_m(H_m) + H_m$, where $\text{MLP}_m$ comprises two linear layers with ReLU activation. The teacher signal is a router-weighted combination of these refined representations:
\begin{equation}
\mathcal{T}_{feat} = \mathrm{StopGradient}\Big( \sum_{m\in\{l,v,a\}} w_m H'_m \Big)
\end{equation}
Since $\mathcal{L}_{mcm}$ optimizes $w_m$, $\mathcal{T}_{feat}$ constitutes a complementarity-aware feature combination where low-weight (noisy) modalities are naturally attenuated. We align fused features with this teacher by minimizing distributional discrepancy:
\begin{equation}
\mathcal{L}_{distill}
= \frac{1}{D} \sum_{j=1}^{D}
\big(
\mathrm{Softmax}(O'_{multi})_j
-
\mathrm{Softmax}(\mathcal{T}_{feat})_j
\big)^2
\end{equation}
This forms a closed loop: the MCM identifies complementarity at the decision level via $w$, while the distillation enforces it at the feature level via $\mathcal{L}_{distill}$.

\subsection{Loss Function}
\label{sec:loss_function}
The total training objective integrates the task loss, complementarity routing loss, and entropy regularization terms.
The task loss aggregates the primary and auxiliary prediction objectives:
\begin{equation}
\mathcal{L}_{task} = \mathcal{L}_{multi} + \mathcal{L}_{uni} + \lambda_{sub} \mathcal{L}_{sub}
\end{equation}
where $\mathcal{L}_{multi}$ is the full-modal primary loss, $\mathcal{L}_{uni} = \frac{1}{3}(\mathcal{L}_l + \mathcal{L}_v + \mathcal{L}_a)$ is the mean unimodal auxiliary loss, and $\mathcal{L}_{sub} = \frac{1}{3}(\mathcal{L}_{va} + \mathcal{L}_{la} + \mathcal{L}_{lv})$ is the MCM-introduced average bimodal branch loss, weighted by $\lambda_{sub}$.
To prevent routing collapse to a single modality or band, we apply entropy regularization to both modality ($\mathbf{w}$) and band ($\alpha_m$) routing weights. The modality-level form is:
\begin{equation}
    \mathcal{L}_{ety} = \frac{N}{B}\sum_{i=1}^{B}\sum_{m=1}^{N} w_m^{(i)} \log w_m^{(i)}
\end{equation}
where $N$ is the number of modalities and $B$ is the batch size. Minimizing $\mathcal{L}_{ety}$ maximizes the entropy of the weight distribution, encouraging balanced modality utilization. The band-level entropy regularization $\mathcal{L}_{band}$ follows the same formulation applied to $\alpha_m$ for each modality. The complementarity routing loss $\mathcal{L}_{mcm}$ and feature distillation loss $\mathcal{L}_{distill}$ are defined in Section~\ref{sec:mcm}.
The final total loss is:
\begin{equation}
\mathcal{L}_{total}
= \mathcal{L}_{task}
+ \lambda_{ety}\mathcal{L}_{ety}
+ \lambda_{mcm}\mathcal{L}_{mcm}
+ \lambda_{dist}\mathcal{L}_{distill}
+ \lambda_{band}\mathcal{L}_{band}
\end{equation}
where $\lambda_{sub}, \lambda_{ety}, \lambda_{mcm}, \lambda_{dist}$, and $\lambda_{band}$ are hyperparameters.

\section{Experiments}
\label{sec:experiments}

To evaluate the generality of our proposed method, we conduct experiments on five public datasets covering both Multimodal Emotion Recognition (MER) and Multimodal Intent Recognition (MIR), including CMU-MOSI~\citep{Zadeh2016MOSI}, CMU-MOSEI~\citep{Zadeh2018MOSEI}, CH-SIMS~\citep{Yu2020CHSIMS}, CH-SIMSv2 \citep{Liu2022CHSIMSv2}, and MIntRec~\citep{Zhang2022MIntRec}. For emotion datasets we report Acc-7 (MOSI/MOSEI), Acc-5/Acc-3 (CH-SIMS/v2), Acc-2, F1, and MAE; for MIntRec we report Accuracy, F1, Precision, and Recall. Implementation details and per-dataset hyperparameters are provided in the supplementary material.

\begin{table*}[htbp]
\centering
\renewcommand{\arraystretch}{1.0}
\setlength{\tabcolsep}{5pt}
\resizebox{\textwidth}{!}{
\begin{tabular}{l|>{\centering\arraybackslash}p{1.7cm}|cccc|cccc}
\toprule
\multirow{2}{*}{\textbf{Methods}} & \multirow{2}{*}{\textbf{Setting}} & \multicolumn{4}{c}{\textbf{MOSI}} & \multicolumn{4}{c}{\textbf{MOSEI}} \\
 &  & ACC7 $\uparrow$ & ACC2 $\uparrow$ & F1 $\uparrow$ & MAE $\downarrow$ & ACC7 $\uparrow$ & ACC2 $\uparrow$ & F1 $\uparrow$ & MAE $\downarrow$ \\
\midrule
EF-LSTM \citep{Zadeh2017TFN} & \multirow{13}{*}{Aligned} & 33.7 & 75.3 & 75.2 & 1.386 & 47.4 & 78.2 & 77.9 & 0.620 \\
LF-DNN \citep{Zadeh2017TFN} &  & 31.5 & 78.4 & 78.3 & 0.972 & 51.7 & 83.5 & 83.1 & 0.568 \\
TFN \citep{Zadeh2017TFN} &  & 31.9 & 78.8 & 78.9 & 0.953 & 50.9 & 80.4 & 80.7 & 0.574 \\
LMF \citep{Liu2018LMF} &  & 36.9 & 78.7 & 78.7 & 0.931 & 52.3 & 84.7 & 84.5 & 0.564 \\
MFN \citep{Zadeh2018MFN} &  & 35.6 & 78.4 & 78.4 & 0.964 & 50.8 & 84.0 & 84.0 & 0.574 \\
Graph-MFN \citep{Zadeh2018MOSEI} &  & 31.5 & 78.1 & 78.1 & 0.970 & 51.6 & 84.6 & 84.5 & 0.553 \\
MulT \citep{Tsai2019MulT} &  & 35.1 & 80.0 & 80.1 & 0.936 & 52.3 & 82.7 & 82.8 & 0.572 \\
MISA \citep{Hazarika2020MISA} &  & 41.8 & 84.2 & 84.2 & 0.754 & 52.3 & 85.3 & 85.1 & 0.543 \\
Self-MM \citep{Yu2021SelfMM} &  & 45.3 & 84.9 & 84.9 & 0.738 & 53.2 & 84.5 & 84.3 & 0.540 \\
MMIM \citep{Han2021MMIM} &  & 45.8 & 84.6 & 84.5 & 0.717 & 50.1 & 83.6 & 83.5 & 0.580 \\
DMD \citep{Li2023DMD} &  & 46.2 & 83.2 & 83.2 & 0.721 & 52.4 & 84.8 & 84.7 & 0.546 \\
EMOE \citep{Fang2025EMOE} &  & 47.7 & 85.4 & 85.4 & 0.710 & 54.1 & 85.3 & 85.3 & 0.536 \\
\textbf{Atsuko(Ours)} &  & \textbf{48.7} & \textbf{86.3} & \textbf{86.2} & \textbf{0.696} & \textbf{54.4} & \textbf{85.6} & \textbf{85.7} & \textbf{0.530} \\
\midrule
EF-LSTM \citep{Zadeh2017TFN} & \multirow{14}{*}{Unaligned} & 31.0 & 73.6 & 74.5 & 1.420 & 46.3 & 76.1 & 75.9 & 0.594 \\
LF-DNN \citep{Zadeh2017TFN} &  & 32.5 & 78.2 & 78.3 & 0.987 & 52.3 & 83.7 & 83.2 & 0.561 \\
TFN \citep{Zadeh2017TFN} &  & 35.3 & 76.5 & 76.6 & 0.995 & 50.2 & 84.2 & 84.0 & 0.573 \\
LMF \citep{Liu2018LMF} &  & 31.1 & 79.1 & 79.1 & 0.963 & 51.9 & 83.8 & 83.9 & 0.565 \\
MFN \citep{Zadeh2018MFN} &  & 34.7 & 80.0 & 80.1 & 0.971 & 51.3 & 83.2 & 83.3 & 0.567 \\
Graph-MFN \citep{Zadeh2018MOSEI} &  & 34.4 & 79.4 & 79.2 & 0.930 & 51.8 & 84.2 & 84.2 & 0.568 \\
MulT \citep{Tsai2019MulT} &  & 33.2 & 80.3 & 80.3 & 0.933 & 53.2 & 84.0 & 84.0 & 0.556 \\
MISA \citep{Hazarika2020MISA} &  & 43.6 & 83.8 & 83.9 & 0.742 & 51.0 & 84.8 & 84.8 & 0.557 \\
Self-MM \citep{Yu2021SelfMM} &  & 45.7 & 83.4 & 83.6 & 0.724 & 52.9 & 85.3 & 84.8 & 0.535 \\
MMIM \citep{Han2021MMIM} &  & 45.9 & 83.4 & 83.4 & 0.777 & 52.6 & 81.5 & 81.3 & 0.578 \\
DMD \citep{Li2023DMD} &  & 46.7 & 84.0 & 84.0 & 0.721 & 53.1 & 84.7 & 84.7 & 0.536 \\
EMOE \citep{Fang2025EMOE} &  & 47.8 & 85.4 & 85.3 & 0.697 & 53.9 & 85.5 & 85.5 & \textbf{0.530} \\
\textbf{Atsuko(Ours)} &  & \textbf{49.4} & \textbf{86.0} & \textbf{85.9} & \textbf{0.691} & \textbf{54.5} & \textbf{85.7} & \textbf{85.6} & \textbf{0.530} \\
\bottomrule
\end{tabular}
}
\caption{Performance comparison on CMU-MOSI and CMU-MOSEI datasets under aligned and unaligned settings. The best results are highlighted in \textbf{bold}. $\uparrow$ indicates higher is better; $\downarrow$ indicates lower is better.}
\label{tab:sota_mosi_mosei}
\end{table*}

\subsection{Comparison with the State-Of-The-Art Methods}
\subsubsection{Results on the English MER Datasets.}
As shown in Table~\ref{tab:sota_mosi_mosei}, Atsuko consistently achieves the best results on both CMU-MOSI and CMU-MOSEI under aligned and unaligned settings. On CMU-MOSI, the advantage is most pronounced in the unaligned setting, where Acc-7 reaches 49.4\%, surpassing EMOE by 1.6 percentage points with an MAE of 0.691. Under the aligned setting, Acc-7 improves by 1.0 point and MAE decreases by 0.014 relative to EMOE. These gains on a dataset characterized by limited samples and complex emotional expressions confirm the effectiveness of our method in fine-grained emotion discrimination. On the larger CMU-MOSEI, Atsuko uniformly surpasses all baselines under both settings, with the unaligned Acc-7 reaching 54.5\% and MAE attaining 0.530. Given that MOSEI features inconsistent modality quality and significant inter-sample variance, these results indicate the model effectively identifies valuable per-sample modality contributions, thereby reducing prediction errors in emotion regression.

\subsubsection{Results on the Chinese MER Datasets.}
As shown in Table~\ref{tab:sota_sims}, Atsuko achieves the best results on both Chinese datasets. On CH-SIMS, the model surpasses all baselines across every metric, with Acc-5 improving by 2.19 percentage points over the second-best method EMOE, confirming its capability in processing Chinese multimodal emotional semantics. On the larger CH-SIMSv2, Atsuko attains the highest Acc-5, Acc-2, F1, and average scores despite slightly trailing BBFN on Acc-3, with the overall average improving by 0.55 percentage points over EMOE. These consistent results validate the cross-lingual generalization and stability of the proposed method in large-scale scenarios.

\begin{table}[thbp]
\centering
\renewcommand{\arraystretch}{1.0}
\setlength{\tabcolsep}{4pt}
\resizebox{\linewidth}{!}{
\begin{tabular}{l|ccccc|ccccc}
\toprule
\multirow{2}{*}{\textbf{Methods}} & \multicolumn{5}{c|}{\textbf{SIMS}} & \multicolumn{5}{c}{\textbf{SIMS\_v2}} \\
 & ACC5 $\uparrow$ & ACC3 $\uparrow$ & ACC2 $\uparrow$ & F1 $\uparrow$ & Avg $\uparrow$ & ACC5 $\uparrow$ & ACC3 $\uparrow$ & ACC2 $\uparrow$ & F1 $\uparrow$ & Avg $\uparrow$\\
\midrule
TFN \citep{Zadeh2017TFN} & 39.30 & 65.12 & 78.38 & 78.62 & 65.36 & 52.55 & 72.21 & 80.14 & 80.14 & 71.26 \\
LMF \citep{Liu2018LMF} & 40.53 & 64.68 & 77.77 & 77.88 & 65.19 & 47.79 & 64.90 & 74.18 & 73.88 & 65.19 \\
Self-MM \citep{Yu2021SelfMM} & 41.53 & 65.47 & 80.04 & 80.44 & 66.87 & 52.77 & 72.61 & 79.69 & 76.76 & 70.46 \\
BBFN \citep{Han2021BBFN} & 40.92 & 61.05 & 78.12 & 77.88 & 64.49 & 53.29 & \textbf{74.47} & 78.53 & 78.41 & 71.18 \\
CENet \citep{Wang2022CENet} & 33.92 & 62.58 & 77.90 & 77.53 & 63.66 & 53.04 & 73.1 & 79.56 & 79.63 & 71.33 \\
EMOE \citep{Fang2025EMOE} & 42.23 & 65.43 & 79.64 & 79.69 & 66.75 & 53.71 & 71.08 & 80.85 & 80.76 & 71.60 \\
\textbf{Atsuko(Ours)} & \textbf{44.42} & \textbf{66.74} & \textbf{80.41} & \textbf{80.21} & \textbf{67.95} & \textbf{54.16} & 72.15 & \textbf{81.17} & \textbf{81.12} & \textbf{72.15} \\
\bottomrule
\end{tabular}
}
\caption{Performance comparison on CH-SIMS and CH-SIMS\_v2 datasets. The best results are highlighted in \textbf{bold}.}
\label{tab:sota_sims}
\end{table}

\subsubsection{Results on the MIR Dataset.}
\begin{wraptable}{r}{0.62\textwidth}
\vspace{-1.2em}
\centering
\renewcommand{\arraystretch}{1.0}
\setlength{\tabcolsep}{4pt}
\small
\begin{tabular}{l|cccc}
\toprule
\textbf{Methods} & ACC $\uparrow$ & F1 $\uparrow$ & P $\uparrow$ & R $\uparrow$ \\
\midrule
MAG-BERT \citep{Rahman2020MAG} & 70.34 & 68.19 & 68.31 & 69.36 \\
MulT \citep{Tsai2019MulT} & 72.58 & 69.36 & 70.73 & 69.47 \\
MISA \citep{Hazarika2020MISA} & 72.36 & 70.57 & 71.24 & 70.41 \\
EMOE \citep{Fang2025EMOE} & 72.58 & 70.73 & 72.08 & \textbf{70.86} \\
\textbf{Atsuko (Ours)} & \textbf{74.16} & \textbf{71.22} & \textbf{72.93} & \textbf{70.86} \\
\bottomrule
\end{tabular}
\caption{Results on MIntRec dataset.}
\label{tab:sota_mintrec}
\vspace{-0.5em}
\end{wraptable}
To validate cross-task generality, we evaluate Atsuko on the multimodal intent recognition dataset MIntRec. As shown in Table~\ref{tab:sota_mintrec}, Atsuko achieves the highest accuracy (74.16\%) and F1 score (71.22\%), surpassing all baselines. Compared to EMOE, Atsuko improves precision from 72.08\% to 72.93\% while matching its recall, indicating that complementarity modeling sharpens the decision boundary. Since MIntRec comprises 20 fine-grained intent categories with an imbalanced distribution, these results confirm that routing captures discriminative cues in long-tail categories and generalizes effectively beyond emotion recognition.

\subsection{Ablation Study}
To further validate the effectiveness of the proposed method, we conduct comprehensive ablation studies on the MOSI and MOSEI datasets.

\begin{table}[htbp]
\captionsetup{skip=1em}
\centering
\renewcommand{\arraystretch}{1.0}
\setlength{\tabcolsep}{4pt}
\resizebox{\linewidth}{!}{
\begin{tabular}{ll|cccc|cccc}
\toprule
\multirow{2}{*}{SBN} & \multirow{2}{*}{MCM} & \multicolumn{4}{c|}{\textbf{MOSI}} & \multicolumn{4}{c}{\textbf{MOSEI}} \\
 &  & ACC7 $\uparrow$ & ACC2 $\uparrow$ & F1 $\uparrow$ &MAE $\downarrow$ & ACC7 $\uparrow$ & ACC2 $\uparrow$ & F1 $\uparrow$ & MAE $\downarrow$\\
\midrule
 &  & 46.0 & 83.5 & 83.5 & 0.716 & 52.3 & 83.8 & 83.7 & 0.558 \\
\checkmark &  & 47.1 & 85.5 & 85.5 & 0.701 & 53.0 & 85.1 & 85.2 & 0.548 \\
 & \checkmark & 46.9 & 85.4 & 85.3 & 0.700 & 53.8 & 85.2 & 85.1 & 0.535 \\
\checkmark & \checkmark & 48.7 & 86.3 & 86.2 & 0.696 & 54.4 & 85.6 & 85.7 & 0.530 \\
\bottomrule
\end{tabular}
}
\caption{Ablation study of the key components in Atsuko on MOSI and MOSEI dataset. $\uparrow$ indicates higher is better; $\downarrow$ indicates lower is better.}
\label{tab:ablation}
\end{table}

\subsubsection{Quantitative Analysis.}
Table~\ref{tab:ablation} presents the ablation results on the MOSI and MOSEI datasets. Each proposed module independently improves overall performance, with complementary contributions. The SBN primarily benefits classification metrics: on MOSI, Acc-2 and F1 rise by approximately 2.0 percentage points, while on MOSEI, Acc-2 gains 1.3 points. By decomposing temporal signals into orthogonal frequency bands and applying dynamic weighting, the SBN effectively isolates fine-grained emotional cues distributed across different frequencies. In contrast, the MCM yields larger gains on regression-oriented metrics. On MOSEI, the MCM alone reduces MAE by 0.023, surpassing the SBN-only reduction of 0.010, confirming that explicitly quantifying the marginal contribution of each modality suppresses redundant information and improves intensity-level prediction. Combining both modules, the full Atsuko model achieves the best results across all metrics, indicating a synergistic effect: the SBN supplies high-quality band-decoupled representations at the feature level, while the MCM ensures they are fused according to genuine information gain at the decision level. A detailed modality contribution analysis under different bimodal and trimodal combinations is provided in the supplementary material.

\subsubsection{Sensitivity Analysis.}
\begin{figure}[ht]
    \centering
    \includegraphics[width=\textwidth]{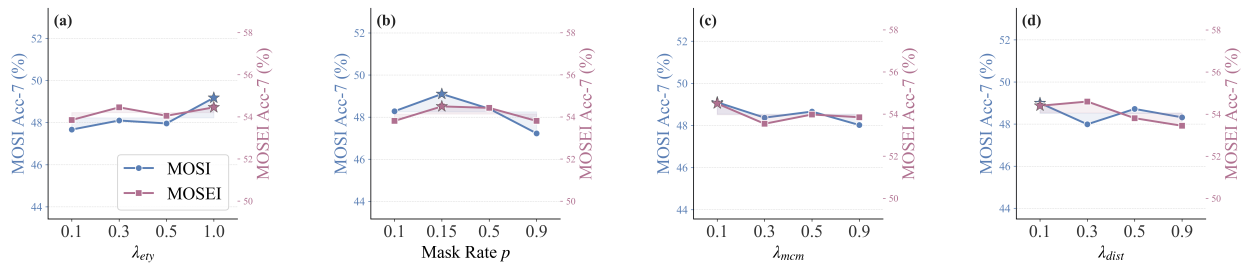}
    \caption{Hyperparameter sensitivity analysis on CMU-MOSI and CMU-MOSEI (Acc-7, \%).
    Each sub-figure sweeps one hyperparameter while holding all others at their default values.
    The star marker ($\bigstar$) denotes the default configuration adopted in all experiments.
    Blue lines: MOSI (left $y$-axis); red lines: MOSEI (right $y$-axis).}
    \label{fig:sensitivity}
\end{figure}

We sweep four key hyperparameters over broad ranges while fixing the others at their defaults: the entropy regularization weight $\lambda_{ety}$, the stochastic mask rate $p$, the complementarity routing loss weight $\lambda_{mcm}$, and the feature distillation loss weight $\lambda_{dist}$. Figure~\ref{fig:sensitivity} reports Acc-7 (\%) on CMU-MOSI and CMU-MOSEI.
For $\lambda_{ety}$ (Figure~\ref{fig:sensitivity}(a)), Acc-7 varies by only 1.51 points on MOSI and 0.60 points on MOSEI across the range [0.1, 1.0], with the default value of 1.0 achieving the best result. For the mask rate $p$ (Figure~\ref{fig:sensitivity}(b)), performance peaks at the default $p{=}0.15$ and degrades gracefully as $p$ increases; even at $p{=}0.9$, MOSI Acc-7 drops by only 1.87 points, confirming that the MCM complementarity estimation tolerates substantial masking noise. For $\lambda_{mcm}$ and $\lambda_{dist}$ (Figures~\ref{fig:sensitivity}(c)--(d)), both achieve their optima at the default value of 0.1; larger values over-constrain the router or shift optimization away from the task loss, yet the total variation remains within 1.08 and 1.01 points on MOSI and 1.15 points on MOSEI.

Overall, the maximum Acc-7 variation across all four hyperparameters is bounded within 1.87 points on MOSI and 1.15 points on MOSEI, confirming that the default configuration generalizes well without per-dataset tuning.

\subsubsection{Visualization of Feature Distribution.}
\begin{figure}[t]
\centering
\includegraphics[width=\linewidth]{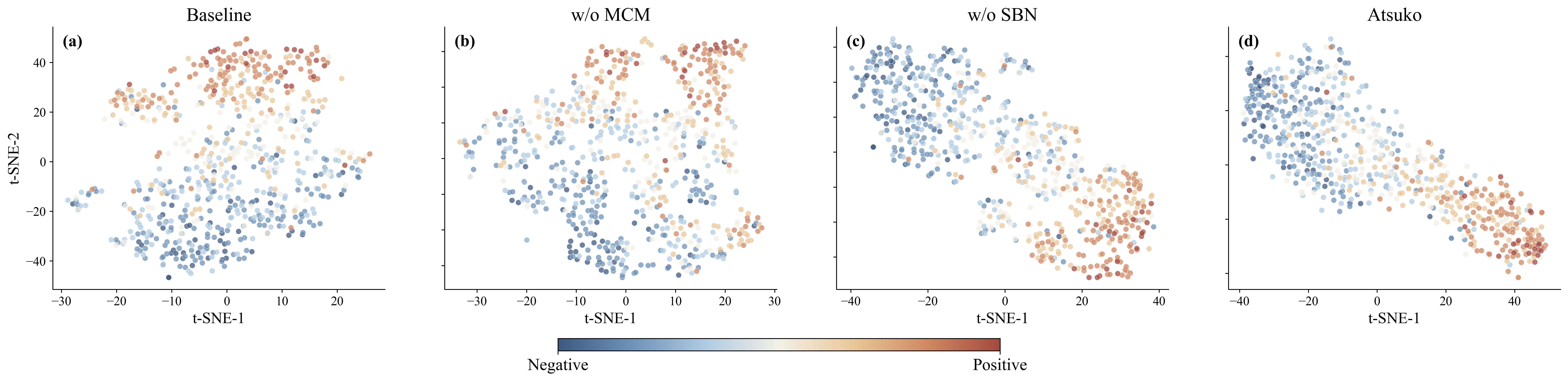}
\caption{t-SNE visualization of feature distributions on MOSI test set. (a) Baseline (w/o SBN \& MCM), (b) w/o MCM, (c) w/o SBN, and (d) Atsuko (Full). Different colors represent varying sentiment intensities. The proposed method shows clearer cluster margins and better separability.}
\label{fig:tsne}
\end{figure}

Figure~\ref{fig:tsne} presents t-SNE visualizations of multimodal features on the MOSI test set under four ablation configurations. The baseline model produces disorganized embeddings with entangled sentiment polarities, confirming that naive concatenation cannot decouple emotional semantics. Introducing either the SBN or the MCM alone yields improved clustering structure, yet inter-class boundaries remain ambiguous. Atsuko model produces the most compact and well-separated clusters, indicating that the SBN filters modality noise while the MCM enhances discriminability via complementarity supervision, and synergy substantially elevates semantic separability.

\subsubsection{Visualization of Marginal Complementarity Module.}
\begin{figure}[t]
\centering
\includegraphics[width=\linewidth]{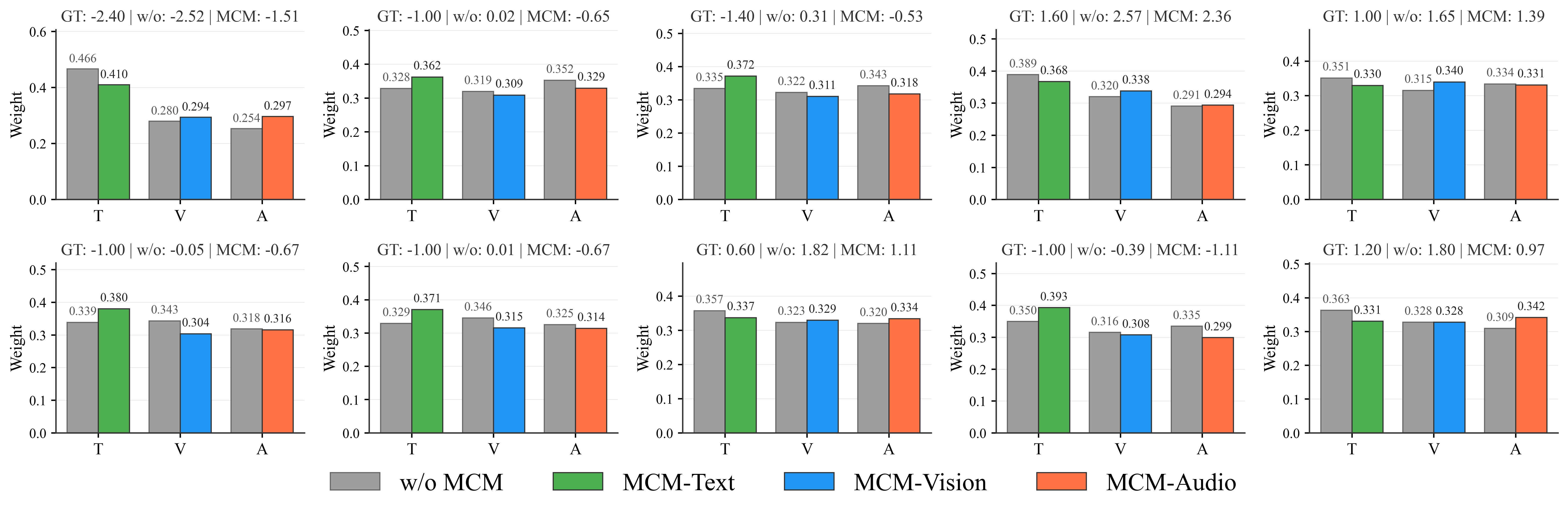}
\caption{Visualization of modality weights predicted by the router on randomly selected CMU-MOSI test samples. Gray bars denote the baseline model without MCM, and colored bars denote the MCM-enhanced model.}
\label{fig:mcm_weights}
\end{figure}
To examine how the MCM influences routing decisions, we visualize modality weights for 10 randomly selected CMU-MOSI test samples in Figure~\ref{fig:mcm_weights}. Two observations emerge. First, for samples where the baseline misclassifies sentiment polarity, the MCM adaptively reallocates weights toward non-dominant modalities, leveraging prosodic and facial cues to correct predictions. This confirms that the MCM suppresses shortcut reliance on a single modality when other modalities carry genuinely complementary cues. Second, when the textual modality does provide the largest marginal contribution, the MCM appropriately increases its weight, demonstrating that the mechanism does not uniformly penalize the dominant modality but rather faithfully reflects per-sample complementarity. Together, these results show that the router distributes attention according to marginal contributions rather than feature confidence alone.

\subsubsection{Visualization of Spectral Band Decomposition.}
To verify whether the SBN captures fine-grained emotional cues, we visualize the spectral band decomposition for a strongly positive sample (Label: +2.0) from the CMU-MOSI test set, showing spectral energy distributions, time-domain band decompositions, and learned router weights across all three modalities.

\begin{figure}[H]
\centering
\begin{subfigure}[b]{\linewidth}
  \centering
  \includegraphics[width=0.95\linewidth]{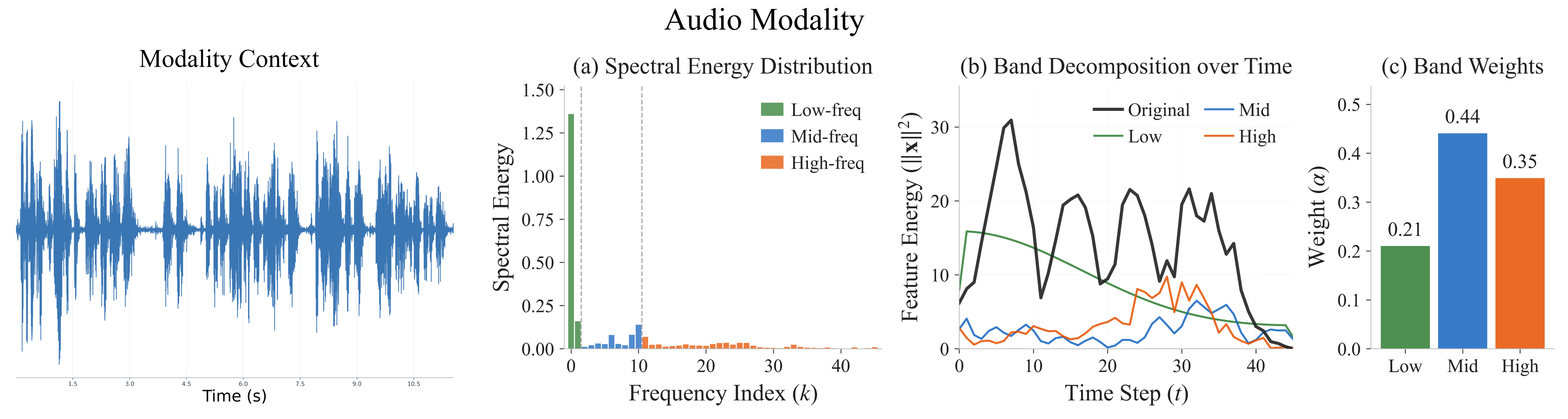}
  \caption{\textbf{Audio} modality. Each panel shows: (a) spectral energy distribution, where colored regions indicate low/mid/high-frequency bands; (b) time-domain feature energy $\|\mathbf{x}\|^2$ decomposed into three bands; (c) learned routing weights $\alpha$. The same panel layout applies to (b) and (c) below.}
  \label{fig:audio}
\end{subfigure}\\[0.3em]
\begin{subfigure}[b]{\linewidth}
  \centering
  \includegraphics[width=0.95\linewidth]{text2.png}
  \caption{\textbf{Text} modality (same sample and panel layout).}
  \label{fig:text}
\end{subfigure}\\[0.3em]
\begin{subfigure}[b]{\linewidth}
  \centering
  \includegraphics[width=0.95\linewidth]{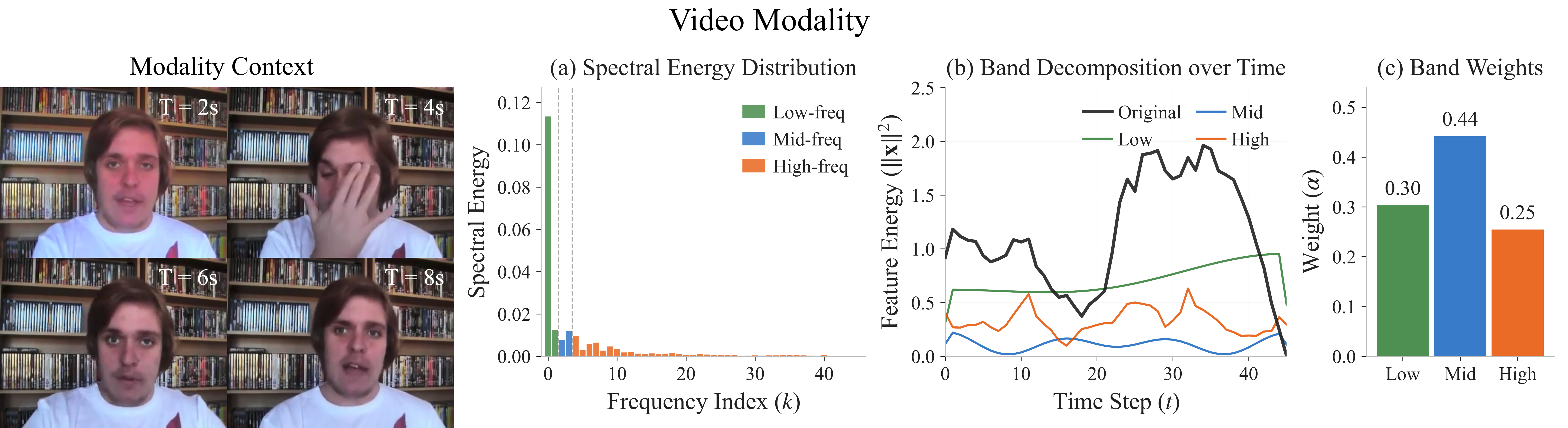}
  \caption{\textbf{Vision} modality (same sample and panel layout).}
  \label{fig:vision}
\end{subfigure}
\caption{Spectral band decomposition of a strongly positive sample (Label: +2.00) from the CMU-MOSI test set across all three modalities.}
\label{fig:spectral_all}
\end{figure}

As shown in Figure~\ref{fig:audio}, Figure~\ref{fig:text}, and Figure~\ref{fig:vision}, the three modalities exhibit distinct spectral structures that the router exploits adaptively. For audio (Figure~\ref{fig:audio}), the low-frequency band shows flat energy corresponding to background noise with limited emotional discriminability, while the mid- and high-frequency bands capture prosodic fluctuations and transient acoustic events such as stress and pauses. The router accordingly assigns dominant weights to mid ($\alpha_{mid} \approx 0.44$) and high ($\alpha_{high} \approx 0.35$) frequencies, suppressing the uninformative low-frequency components. For text (Figure~\ref{fig:text}), energy concentrates in ultra-low frequency bands ($k=0, 1$), reflecting deep semantic features from BERT that encode global contextual information with smooth temporal evolution. The router assigns dominant weight to the low-frequency band ($\alpha_{low} \approx 0.39$), correctly identifying that core emotional semantics reside in stable contextual representations. For vision (Figure~\ref{fig:vision}), the low-frequency band encodes static baselines such as identity and head pose, whereas the mid-frequency band captures subtle facial muscle movements and lip dynamics critical for expression recognition. The router assigns the highest weight to the mid-frequency band ($\alpha_{mid} \approx 0.44$), focusing on dynamic facial regions.
Across modalities, feature energy follows a clear hierarchy (text $>$ audio $>$ vision), reflecting inherent differences in information density: BERT features encapsulate dense semantic information per time step, acoustic features combine continuity with discrete prosodic variations, and visual features exhibit high inter-frame redundancy. Through spectral band routing, the SBN bridges these order-of-magnitude energy differences to adaptively extract prosodic cues from audio, global semantics from text, and facial dynamics from vision, achieving effective multimodal fusion aligned with human perceptual mechanisms.

\section{Conclusion}
This paper presents Atsuko, a complementarity-supervised multi-band expert network for multimodal emotion recognition that addresses coarse-grained fusion and modality dominance. The Spectral-Band Hierarchical Routing Network extends dynamic routing into the spectral domain, enabling fine-grained decoupling and adaptive selection of emotional cues from heterogeneous signals. The Marginal Complementarity Module incorporates cooperative game theory to explicitly quantify each modality's marginal contribution, enforcing reliance on genuinely complementary evidence and mitigating shortcut learning. Experiments across five benchmarks validate the superior performance, robustness, and interpretability of our method. The proposed spectral-temporal routing paradigm offers a versatile fusion framework for finer-grained representation decoupling in affective computing.

\bibliographystyle{unsrtnat}
\bibliography{arxiv}

\end{document}